\documentclass[conference,a4paper]{APSIPA2025}
\usepackage{cite}
\usepackage{url}
\usepackage{booktabs}
\usepackage{amsmath,amssymb,amsfonts}
\usepackage{graphicx}
\usepackage{multirow}
\usepackage{threeparttable}
\usepackage{algorithmic}
\usepackage{graphicx}
\usepackage{textcomp}
\usepackage{tabularx}
\usepackage{xcolor}
\usepackage{hyperref}
\usepackage{makecell}

\usepackage{geometry}
\usepackage{fancyhdr}

\fancypagestyle{firststyle}{
  \fancyhf{}
  \fancyhead[C]{2025 Asia Pacific Signal and Information Processing Association Annual Summit and Conference (APSIPA ASC)}
}

\begin{document}

\title{Multi-task Pretraining for Enhancing Interpretable L2 Pronunciation Assessment\\
\thanks{Two images made by soco-st
was used in this paper. \url{https://soco-st.com}}
}

\author{
\authorblockN{
Jiun-Ting Li\authorrefmark{1},
Bi-Cheng Yan\authorrefmark{2},
Yi-Cheng Wang\authorrefmark{3},
Berlin Chen\authorrefmark{2},
}

\authorblockA{
\authorrefmark{1}
Advanced Technology Laboratory, Chunghwa Telecom Co., Ltd., Taiwan}

\authorblockA{
\authorrefmark{2}
National Taiwan Normal University, Taiwan}

\authorblockA{
\authorrefmark{3}
National Taiwan University, Taiwan}
}


\maketitle
\thispagestyle{firststyle}
\pagestyle{fancy}


\begin{abstract}
Automatic pronunciation assessment (APA) analyzes second-language (L2) learners’ speech by providing fine-grained pronunciation feedback at various linguistic levels. Most existing efforts on APA typically adopt segmental-level features as inputs and predict pronunciation scores at different granularities via hierarchical (or parallel) pronunciation modeling. This, however, inevitably causes assessments across linguistic levels (e.g., phone, word, and utterance) to rely solely on phoneme-level pronunciation features, nearly sidelining supra-segmental pronunciation cues. To address this limitation, we introduce multi-task pretraining (MTP) for APA, a simple yet effective strategy that attempts to capture long-term temporal pronunciation cues while strengthening the intrinsic structures within an utterance via the objective of reconstructing input features. Specifically, for a phoneme-level encoder of an APA model, the proposed MTP strategy randomly masks segmental-level pronunciation features and reconstructs the masked ones based on their surrounding pronunciation context. Furthermore, current APA systems lack integration with automated speaking assessment (ASA), limiting holistic proficiency evaluation. Drawing on empirical studies and prior knowledge in ASA, our framework bridges this gap by incorporating handcrafted features (HCFs), such as fluency (speech rate, silence duration) and stress (pitch accent strength), derived from human-designed formulas via regressors to generate interpretable proficiency scores. Experiments on speechocean762 show improved pronunciation scoring and ASA proficiency correlation, enabling targeted training and comprehensive proficiency assessment.
\end{abstract}

\section{Introduction}
\label{sec:intro}

Fueled by frequent global economic and cultural exchanges, foreign language acquisition, such as English, has become increasingly vital \cite{PHowson_2013_the_english_effect}.  However, the insufficient supply of language instructors struggles to keep pace with the growing needs of the second language (L2) learners. In response, computer-assisted language learning (CALL) emerges as a promising solution to bridge this gap, which provides effective self-directed learning environments through automatic scoring systems that ensure greater consistency and speed at a lower cost. \cite{zhang2013contrasting, doi:10.1177/0033688220977406}. As a core component of CALL, automatic pronunciation assessment (APA) aims to evaluate the oral skills of L2 learners by analyzing their pronunciation quality in terms of the presented text prompts (or reference texts) in a target language \cite{kheir_etal_2023_automatic}.

\begin{figure}[h!]
  \centering
  \includegraphics[width=\linewidth]{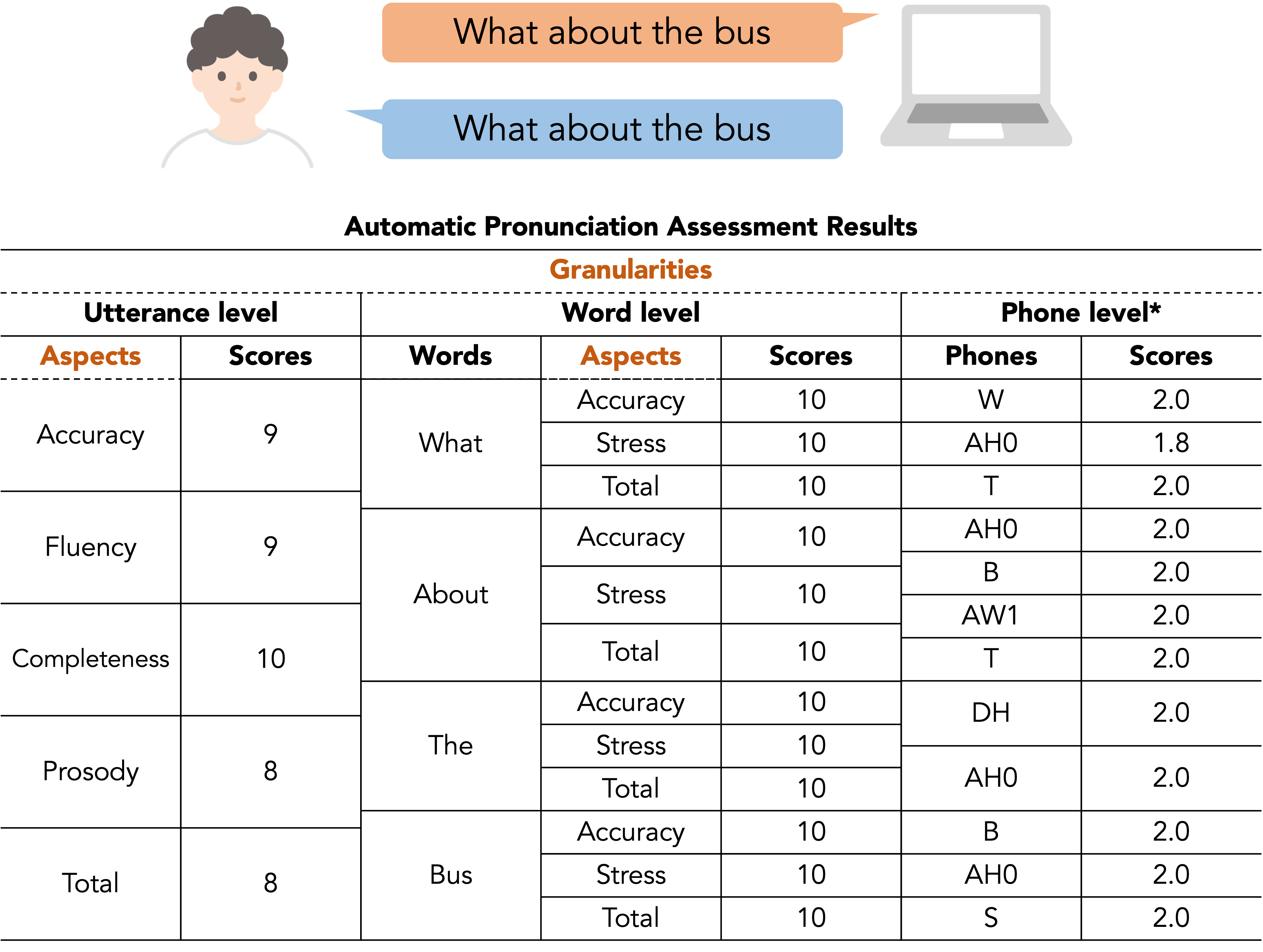}
  \caption{An illustration of a partial pronunciation assessment sample in the speechocean762 \cite{zhang21x_interspeech} corpus, demonstrating multi-granularity evaluation at the phone, word, and utterance levels, with multi-aspect pronunciation metrics applied across each granularity to support comprehensive L2 feedback. $^{*}$ indicates the single aspect called accuracy at the phoneme-level (granularity).}
  \label{fig:pronunciation_assessment}
  \vspace{-0.6cm}
\end{figure}

As depicted in Figure \ref{fig:pronunciation_assessment}, an L2 learner is presented with a reference text, and in turn, the APA system analyzes their speech alongside the reference text by assessing their pronunciation proficiency with fine-grained pronunciation aspects (e.g., accuracy, stress, and fluency) across multiple linguistic levels (viz., phoneme, word, and utterance levels) \cite{9746743, 10890778}. 

\begin{figure*}[h!]
  \centering
  \includegraphics[width=\linewidth]{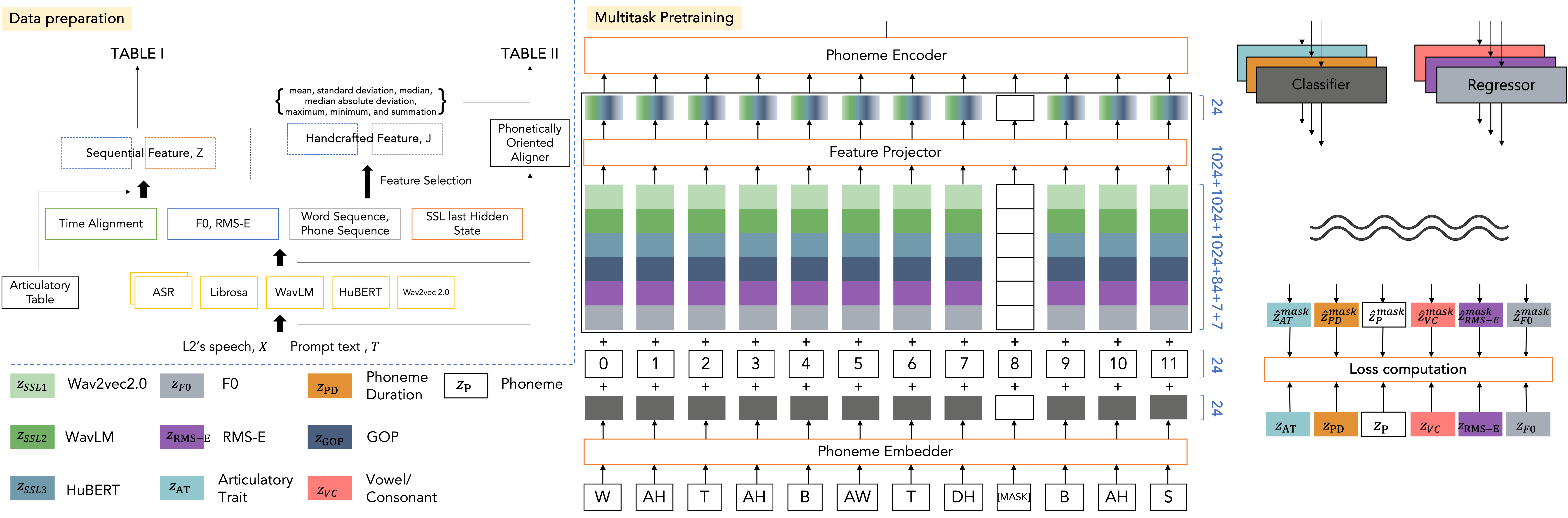}
  \caption{An illustration of the data preparation process and the MTP framework. In the data preparation process, L2 speech and prompt text are processed by pre-trained acoustic models to extract time alignment information, deep features, and acoustic features (e.g., F0 contours, RMS energy). These are transformed into sequential features and aggregated into handcrafted features using human-designed formulas, supporting MTP and APA training.}
  \label{fig.pretrain_multitask_framework}
  \vspace{-0.5cm}
\end{figure*}

The existing studies on APA typically employ a unified pronunciation assessment model with a parallel \cite{9746743, chao20223m} or hierarchical \cite{10890778, chao23_interspeech, 10095733} neural architecture to assess pronunciation aspects across linguistic granularities by leveraging segmental-level pronunciation features (e.g., the statistics of energy and duration, the features of phoneme-level pronunciation quality and textual information) in conjunction with a feature aggregation mechanism. A building block for these models, goodness of pronunciation (GOP) feature measures the pronunciation deviation between L2 learners and native speakers at the phoneme-level \cite{9746743}. It first utilized GOP features as the input for APA systems, derived from an automatic speech recognizer (ASR). While GOP features remain in use, recent APA models increasingly leverage deep learning paradigms, such as self-supervised learning (SSL) models \cite{10890778, chao20223m, chao23_interspeech}, to just name a few, Wav2vec2.0 \cite{baevski2020wav2vec}, HuBERT \cite{hsu2021hubert} and WavLM \cite{9814838}, to automatically extract contextual features. Pretrained on large-scale unlabeled speech data, these SSL models capture phonological patterns in English, mitigating phonetic variability and enhancing scalability and phoneme-level precision. This evolution opens new avenues for advancing pronunciation assessment.

Albeit models stemming from the effective modeling paradigm have demonstrated promising results on a few APA tasks, they still suffer from at least two weaknesses. First, the assessments across linguistic levels (e.g., phone, word, and utterance) rely solely on segmental-level pronunciation features, which inevitably sideline the supra-segmental pronunciation cues. While recent APA models, while leveraging several features for overall precision, such as phonetic features (e.g., phoneme sequence), prosodic features (e.g., energy, pitch and phoneme duration), to create multi-view APA \cite{10890778, chao20223m, 10389777}, they do not jointly model phonetic and prosodic content and rely heavily on predictors, with encoders failing to intrinsically encode phonetic and prosody-aware representations. To address the limitations of accumulating features without optimization, and treating features solely as inputs, we propose multi-task pretraining (MTP) for APA, which optimizes the phoneme encoder to integrate sequential phonetic and prosodic features as inputs and outputs within hierarchical architectures \cite{10890778, 10389777}, reconstructing them to enhance APA. Our framework incorporates phonetic subtasks, including phoneme prediction \cite{jia21_interspeech, thenguyen23_interspeech, fu23_interspeech}, articulatory traits prediction, and vowel/consonant prediction, alongside prosodic subtasks like phoneme duration prediction \cite{fu23_interspeech}, pitch prediction and energy prediction aggregated from phoneme alignment information, representing fundamental frequency (F0), and root-mean-square energy (RMS-E), respectively.

\begin{table}[t!]
    \centering
    \caption{Task details with views, denotations, and dimension size}
    \label{tab:task_details}
    \begin{tabular}{llcc}
    \hline
    \textbf{Tasks}   & \textbf{Views}                       & \textbf{Denotations} & \textbf{Dimension size} \\ \hline
    \multirow{3}{*}{Phonetic} 
                     & Articulation Trait                   & $S_{A}$             & 1                       \\ \cline{2-4} 
                     & Phoneme Sequence                       & $S_P$                & 1                       \\ \cline{2-4} 
                     & Vowel/Consonant                      & $S_V$                & 1                       \\ \hline  
    \multirow{3}{*}{Prosodic} 
                     & RMS-E Statics                               & $S_{R}$          & 7                       \\ \cline{2-4} 
                     & F0 Statics                                  & $S_{F}$             & 7                       \\ \cline{2-4} 
                     & Phoneme Duration                       & $S_{D}$             & 1                       \\ \hline
    \end{tabular}
\vspace{-0.5cm}
\end{table}

Second is the lack of connection from APA to automated speaking assessment (ASA), hindering comprehensive proficiency evaluation. We propose to integrate the handcrafted features (HCFs) from automated speaking assessment (ASA) into APA, which exploit alignment information and follow human-designed formulas to compute, including fluency \cite{strik1999automatic, preciado2018speaker}, pronunciation \cite{10.1007/978-3-030-50729-9_48, 5700836}, and rhythm \cite{kyriakopoulos19_interspeech}, aggregated across utterance-level for offering interpretable L2 scores \cite{7820782, zechner2019automated}, e.g., speech rate too slow. Our approaches pave the way for more holistic and precise APA systems. To summarize the contributions in this work: (1) a MTP framework that infuses phonetic and prosodic features into the phoneme encoder for promoting the overall APA model's performance; and (2) the HCFs in ASA are integrated into the APA, paving the way to a comprehensive system.

\section{Methodology}
\label{sec:methodology}

\subsection{Task Formulation and Model Architecture}
\label{subsec:task_formulation_and_model_architecture}

We tackle the APA task with the following flow. Given $X$ is the L2 learner’s speech input and $T$ is the reference text prompt, the goal is to produce a proficiency score vector $\mathbf{s}$, encompassing aggregate and multi-aspect scores at phone, word, and utterance levels. Formally, APA is defined as a mapping $\mathbf{s} = f(X, T; \theta)$, where $f$ is a model parameterized by $\theta$, evaluating pronunciation by comparing $X$ to $T$ \cite{kheir_etal_2023_automatic} as depicted in Figure~\ref{fig.pretrain_multitask_framework}, the preprocessing extracts phonetic and prosodic sequential features $\text{Z}$ (e.g., phoneme alignments, duration, pitch) from $X$ and $T$, and the fine-grained HCFs $\text{J}$, which are used to generate multi-aspect proficiency scores.

The APA model comprises encoders and scoring modules. The phoneme encoder, denoted as $\theta_p$, processes $\text{Z}$ to produce a sequence of latent representations $\text{H} = \{\mathbf{h_1}, \mathbf{h_2}, \dots, \mathbf{h_n}\}$, where each $\mathbf{h_i}$ encodes phoneme-level features. We adopt a hierarchical architecture, inspired by \cite{10890778}, to encode multi-granularity (phone, word, utterance) representations, using an E-Branchformer as the backbone \cite{10022656}, integrating convolutional and Branchformer layers to capture local and global dependencies, enhancing each level's precision. The scoring modules, denoted as $\theta_s$, receive the projected $\text{J}$ and map $\text{H}$ to proficiency scores $\mathbf{s}$, including aggregate scores (e.g. total) and multi-aspect metrics (e.g., pronunciation accuracy, prosody quality), via regression heads for each granularity level. To optimize $\theta_p$, we employ the MTP method to promote overall performance.

\subsection{MTP Framework with Phonetic and Prosodic Subtasks}
\label{subsec:mtp_framework_with_phonetic_and_prosodic_subtasks}

\begin{table}[t]
\caption{Delivery is an important aspect of pronunciation assessment, focusing on the analysis and use of acoustic information. It is divided into pronunciation and fluency, along with the feature elements of granularity listed below \cite{kyriakopoulos2021deep, zechner2019automated}.}
\label{tab:asa_attributes}
\centering
\begin{tabular}{cccl}
\hline
\textbf{Evaluation} & \textbf{Aspects} & \textbf{Features} & \textbf{Items} \\ \hline
\multirow{22}{*}{\begin{tabular}[c]{@{}c@{}}Delivery\end{tabular}} &
  \multirow{15}{*}{\begin{tabular}[c]{@{}c@{}}Pronunciation\end{tabular}} &
  \multirow{1}{*}{Phone} & Confidence \\ \cline{3-4} 
  & & \multirow{1}{*}{Word} & Confidence \\ \cline{3-4} 
  & & Pitch & F0 \\ \cline{3-4} 
  & & Energy & RMS-E \\ \cline{3-4} 
  & & \multirow{7}{*}{Rhythm} & CCI \\
  & & & rPVI \\
  & & & nPVI \\
  & & & Varco \\
  & & & Number of Nucleus \\
  & & & Number of Consonants \\
  & & & Vowel-to-consonant Ratio \\ \cline{3-4} 
  & & \multirow{2}{*}{Error Rate} & Error Rate (ER) \\
  & & & Match Error Rate (MER) \\ \cline{2-4} 
  &
  \multirow{12}{*}{\begin{tabular}[c]{@{}c@{}}Fluency\end{tabular}} &
  \multirow{2}{*}{Silence} &
  Silence \\
  & & & Long Silence \\ \cline{3-4} 
  & & \multirow{3}{*}{phone} & Duration \\
  & & & Count \\
  & & & Frequency \\ \cline{3-4} 
  & & \multirow{7}{*}{Word} & Duration \\
  & & & Character Length \\
  & & & Count \\
  & & & Frequency \\
  & & & Repeat Count \\
  & & & Distinct Count \\
  & & & Number of Filled Pauses \\ \hline
\end{tabular}
\vspace{-0.5cm}
\end{table}

To enhance APA accuracy, we develop a multi-task learning (MTL) framework that pretrains a phoneme encoder to jointly model phonetic and prosodic features, as shown in Figure \ref{fig.pretrain_multitask_framework}. The phoneme encoder, a 3-layer transformer, processes phonetic and prosodic sequential features as inputs and outputs, as detailed in Table \ref{tab:task_details}. Inspired by BERT \cite{devlin_etal_2019_bert}, we apply a masking strategy, randomly masking 15\% of phonetic features, with 90\% replaced by a mask token and 10\% unchanged, to enhance robustness. Three masking methods are used: 1) replacing phonemes with a mask token, 2) zeroing one-dimensional features (e.g., F0 values), and 3) zeroing multi-dimensional prosodic vectors (e.g., energy profiles). A curriculum learning approach progresses from non-masking teacher-forcing training to masking-based training, optimizing multi-granularity feature integration.

\noindent \textbf{Phoneme Prediction.} Predicts phones in the input sequence. This technique has been applied in text-to-speech \cite{jia21_interspeech, thenguyen23_interspeech} and has achieved success, \cite{fu23_interspeech} also applies this task in their framework. It has 42 categories (39 phones, one \verb|[PAD]| padding token, one \verb|[MASK]| masking token, and one \verb|[UNK]| special token).

\noindent \textbf{Vowel/Consonant Prediction.} Classifies phones as vowels or consonants. This idea has been used as input in \cite{chao20223m}. It has vowel, consonant and a trash can category.

\noindent \textbf{Articulation Trait Prediction}. Predicts articulation traits, extending a binary vowel/consonant prediction, containing vowel, stop, affricate, fricative, aspirate, liquid, nasal, semivowel, and a trash can category. These features have been incorporated in the mispronunciation detection and diagnosis field \cite{10097226}, to the best of our knowledge, it is a novel use in the APA domain.

\noindent \textbf{Phoneme Duration Prediction}. Classifies phoneme-level durations, categorizing frame counts from 1 to 100 (capped at 100 for longer durations). \cite{fu23_interspeech} also uses this task.

\noindent \textbf{Pitch Prediction and Energy Prediction}. Predicts phoneme-level fundamental frequency (F0) and root-mean-square energy (RMS-E) vectors. Using \verb|librosa.pyin| \cite{mcfee2015librosa, 6853678} to extract F0 and \verb|librosa.feature.rms| to extract energy from raw speech data, we aggregate by phoneme duration to compute seven metrics: mean, standard deviation, median, median absolute deviation, maximum, minimum, and sum, yielding seven-dimensional feature vectors.

The MTP loss is a weighted sum of subtask losses, where their weights are hyperparameters tuned to balance subtask contributions, ensuring the encoder captures phonetic and prosodic features for multi-view APA.

\subsection{HCFs in Automated Speaking Assessment}
\label{sec:handcrafted_features_in_automated_speaking_assessment}

In ASA, we categorize the HCFs related to pronunciation assessment into two main aspects: pronunciation and fluency. According to the definition in phonetics \cite{wayland2018phonetics}, the features used for these aspects include elements such as duration, speed, volume, and pitch, which can be obtained through duration, energy, and F0, respectively. The detailed HCF items used are shown in Table \ref{tab:asa_attributes}. Some of the listed HCFs include confidence, energy, silence, long silence, duration, and phoneme length, which are further processed statistically to obtain values such as mean, standard deviation, median, median absolute deviation, maximum, minimum, and summation. Additionally, the total duration of the audio file and the duration of the actual spoken content are included. These are iconic components in ASA for evaluating the overall speaking proficiency. These features are conveyed to the projection linear layers, fusing into the final decisions.

\section{Experiments}
\label{sec:experiments}

\begin{table*}[h!]
\centering
\caption{Performance Metrics for Different Methods. HierCB-imp is the result of our implemented version of HierCB. HierCB-bpe-f0 denoted HCBbf. Comp., Acc. indicates Completeness and Accuracy, respectively. \textbf{Bold} and \underline{underline} denote the best and the second-best performance in each aspect, respectively. $\otimes$ symbol indicates the accumulation of subtasks in MTP.}
\label{tab:experimental_results.without_dataaug}
\scalebox{0.94}{
\begin{tabular}{l|cc|cccc|cccccc}
\toprule
\multirow{2}{*}{\textbf{Methods}} & \multicolumn{2}{c}{\textbf{Phoneme Score}} & \multicolumn{4}{c}{\textbf{Word Score (PCC)}} & \multicolumn{6}{c}{\textbf{Utterance Score (PCC)}} \\
\cmidrule(lr){2-3} \cmidrule(lr){4-7} \cmidrule(lr){8-13}
 & MSE $\downarrow$ & PCC $\uparrow$ & Acc. $\uparrow$ & Stress $\uparrow$ & Total $\uparrow$ & Avg. PCC $\uparrow$ & Acc. $\uparrow$ & Comp. $\uparrow$ & Fluency $\uparrow$ & Prosodic $\uparrow$ & Total $\uparrow$ & Avg. PCC $\uparrow$ \\
\midrule
HierCB\cite{10890778} & 0.076 & 0.680 & 0.630 & 0.355 & 0.645 & - & 0.772 & 0.677 & 0.827 & 0.822 & 0.796 & - \\
\midrule \midrule
HierCB-imp & \makecell{0.078 \\ (0.001)} & \makecell{0.660 \\ (0.002)} & \makecell{0.608 \\ (0.011)} & \makecell{0.385 \\ (0.045)} & \makecell{0.625 \\ (0.011)} & \makecell{0.500 \\ (0.020)} & \makecell{0.757 \\ (0.006)} & \makecell{0.685 \\ (0.153)} & \makecell{0.835 \\ (0.004)} & \makecell{0.826 \\ (0.003)} & \makecell{0.787 \\ (0.004)} & \makecell{0.723 \\ (0.033)} \\
\midrule 
HierCB-f0 & \makecell{0.077 \\ (0.000)} & \makecell{0.665 \\ (0.001)} & \makecell{0.616 \\ (0.006)} & \makecell{0.375 \\ (0.025)} & \makecell{0.635 \\ (0.006)} & \makecell{0.542 \\ (0.012)} & \makecell{0.758 \\ (0.008)} & \makecell{0.836 \\ (0.129)} & \makecell{0.836 \\ (0.007)} & \makecell{0.826 \\ (0.005)} & \makecell{0.787 \\ (0.007)} & \makecell{0.758 \\ (0.031)}\\
\midrule 
HCBbf & \makecell{0.080 \\ (0.000)} & \makecell{0.651 \\ (0.003)} & \makecell{0.602 \\ (0.004)} & \makecell{0.397 \\ (0.033)} & \makecell{0.618 \\ (0.004)} & \makecell{0.539 \\ (0.014)} & \makecell{0.753 \\ (0.007)} & \makecell{0.705 \\ (0.106)} & \makecell{0.833 \\ (0.005)} & \makecell{0.824 \\ (0.004)} & \makecell{0.782 \\ (0.004)} & \makecell{0.779 \\ (0.025)} \\
\hspace{2px}+$S_{P}$ & \makecell{\textbf{0.077} \\ (0.001)} & \makecell{0.670 \\ (0.004)} & \makecell{\textbf{0.631} \\ (0.004)} & \makecell{\underline{0.402} \\ (0.037)} & \makecell{\textbf{0.647} \\ (0.004)} & \makecell{\textbf{0.560} \\ (0.015)} & \makecell{\textbf{0.772} \\ (0.004)} & \makecell{\textbf{0.791} \\ (0.101)} & \makecell{\underline{0.836} \\ (0.004)} & \makecell{\underline{0.826} \\ (0.003)} & \makecell{\textbf{0.796} \\ (0.004)} & \makecell{\textbf{0.804}
 \\ (0.024)} \\
\hspace{4px}$\otimes S_{D}$\cite{fu23_interspeech} & \makecell{0.078 \\ (0.000)} & \makecell{0.659 \\ (0.002)} & \makecell{0.622 \\ (0.002)} & \makecell{0.361 \\ (0.026)} & \makecell{\underline{0.637} \\ (0.022)} & \makecell{\underline{0.540} \\ (0.010)} & \makecell{0.762 \\ (0.002)} & \makecell{0.755 \\ (0.099)} & \makecell{\underline{0.835} \\ (0.004)} & \makecell{\underline{0.824} \\ (0.003)} & \makecell{0.788 \\ (0.002)} & \makecell{\underline{0.793} \\ (0.022)} \\
\hspace{6px}$\otimes S_{V}$ & \makecell{0.078 \\ (0.001)} & \makecell{0.659 \\ (0.002)} & \makecell{0.620 \\ (0.008)} & \makecell{0.316 \\ (0.022)} & \makecell{\underline{0.637} \\ (0.009)} & \makecell{0.524 \\ (0.013)} & \makecell{0.769 \\ (0.002)} & \makecell{0.694 \\ (0.124)} & \makecell{\underline{0.833} \\ (0.008)} & \makecell{\underline{0.826} \\ (0.005)} & \makecell{0.794 \\ (0.002)} & \makecell{\underline{0.783} \\ (0.028)} \\
\hspace{8px}$\otimes S_{A}$ & \makecell{0.079 \\ (0.001)} & \makecell{0.659 \\ (0.002)} & \makecell{0.617 \\ (0.005)} & \makecell{\textbf{0.425} \\ (0.039)} & \makecell{\underline{0.634} \\ (0.004)} & \makecell{\underline{0.559} \\ (0.016)} & \makecell{0.764 \\ (0.003)} & \makecell{0.542 \\ (0.191)} & \makecell{\underline{0.834} \\ (0.003)} & \makecell{\underline{0.824} \\ (0.003)} & \makecell{0.789 \\ (0.002)} & \makecell{0.751 \\ (0.041)} \\
\hspace{10px}$\otimes S_{F}$ & \makecell{0.078 \\ (0.001)} & \makecell{0.660 \\ (0.003)} & \makecell{0.623 \\ (0.004)} & \makecell{0.370 \\ (0.017)} & \makecell{\underline{0.639} \\ (0.003)} & \makecell{\underline{0.544} \\ (0.008)} & \makecell{0.766 \\ (0.002)} & \makecell{0.709 \\ (0.116)} & \makecell{\textbf{0.839} \\ (0.001)} & \makecell{\textbf{0.827} \\ (0.002)} & \makecell{0.790 \\ (0.002)} & \makecell{\underline{0.786} \\ (0.025)} \\
\hspace{12px}$\otimes S_{R}$ & \makecell{0.080 \\ (0.001)} & \makecell{0.649 \\ (0.003)} & \makecell{0.598 \\ (0.013)} & \makecell{\underline{0.400} \\ (0.041)} & \makecell{0.617 \\ (0.011)} & \makecell{0.538 \\ (0.021)} & \makecell{0.755 \\ (0.010)} & \makecell{0.504 \\ (0.234)} & \makecell{0.832 \\ (0.007)} & \makecell{0.822 \\ (0.006)} & \makecell{0.780 \\ (0.008)} & \makecell{0.739 \\ (0.053)} \\
\bottomrule
\end{tabular}
}
\vspace{-0.5cm}
\end{table*}

\subsection{Experimental Setup}
\label{subsec:experimental_setup}

We adhere to the hyperparameters from \cite{10890778} for MTP and training of our APA model. To quantify epistemic uncertainty, we train with five different random seeds. To evaluate the APA model's performance, the Pearson correlation coefficient (PCC) is adopted, while the mean squared error (MSE) is only used for the phoneme accuracy.

To address the skewed word distribution and limited vocabulary of word inputs for the APA model, we employ modernBERT \cite{warner2024smarterbetterfasterlonger}, a BERT variant optimized for robust tokenization, to generate token-level embeddings, replacing the inherent word embeddings in current APA models. To ensure accurate phone-word alignment, we use the Myers diff algorithm \cite{myers1986nd}, which efficiently aligns token sequences to their corresponding words. Multiple token embeddings are then aggregated into a single word embedding using attention pooling, where a clipped softmax of the attention matrix weights the contributions of each token.
For preparing the HCFs in Table \ref{tab:asa_attributes}, we adopt the pretrained ASR model with TDNN-F framework to retrieve the alignment information \cite{zhang21x_interspeech}. But for the error rates (ERs), we adopt an ASR under the framework of a SSL acoustic model with differentiable weighted finite-state transducers \cite{10832327}, pretrained on Librispeech \cite{7178964} corpus, a 960-hour training set, to produce the transcriptions. The transcribed results may deviate from the prompts due to the listener's perceptions of L2 pronunciation errors. Thus, we adopt a phonetically oriented aligner \cite{7404808}, which retains the phonetic and linguistic relationships, to compute the ERs in reference and hypothesis transcripts, including the case of homophonic errors. After extracting HCFs, we eliminate those with skewed distributions, binary features exceeding a threshold (0.8), and duplicates. We then apply feature selection using a multivariate regressor model \cite{tibshirani1996regression} to identify the most relevant HCFs for building an optimized utterance-level proficiency assessment model. Finally, we obtained 56, 66, 51, 50, and 53 dimensions of HCFs for each aspect at the utterance level, totalling 112 unique dimensions.


\subsection{Corpus}
\label{subsec:corpus}

We conducted APA experiments on the speechocean762 corpus \cite{zhang21x_interspeech}, which is a publicly available dataset specifically designed for research on APA [27]. This dataset contains 5,000 English-speaking recordings spoken by 250 Mandarin L2 learners. The training and test sets are of equal size, and each of them has 2,500 utterances, where annotates utterance-level scores (accuracy, stress, completeness, fluency, prosodic, total), word-level scores (accuracy, stress, total) and phoneme accuracy score.

\subsection{Experimental Results}
\label{subsec:experimental_results}

Table \ref{tab:experimental_results.without_dataaug} presents the performance of our APA models compared to the original HierCB model \cite{10890778}. Our reproduced baseline, HierCB-imp replicates HierCB by rerunning the training without altering the implementation, resulting in minor performance differences. We introduce HierCB-f0, which incorporates fundamental frequency (F0) as an additional input feature, and HierCB-bpe-f0 (HCBbf), which further replaces word embeddings with token-level embeddings from modernBERT \cite{warner2024smarterbetterfasterlonger}. Additional MTP subtasks are evaluated, including phoneme prediction ($S_{P}$), duration prediction ($S_{D}$) \cite{fu23_interspeech}, vowel/consonant classification ($S_{V}$), articulation prediction ($S_{A}$), F0 prediction ($S_{F}$), and RMS-E prediction ($S_{R}$).

The results show that HierCB-f0 outperforms HierCB-imp in many aspects, while HCBbf underperforms slightly compared to HierCB-f0 but HCBbf addresses the issue of word embedding; we prefer to use this one in our further experiments.

Consequently, we adopt HCBbf using MTP. The phoneme-, word-, and utterance-level averaged performance peaks with $S_P$, achieving MSE 0.077, PCCs 0.560 and 0.804, respectively. Utterance-level fluency and prosodic metrics peak with $S_F$, reaching PCCs of 0.839 (fluency) and 0.827 (prosodic), outperforming baselines. Adding $S_A$ and $S_R$ subtasks further enhances word-level stress (PCC 0.425, 0.400). These findings suggest that phoneme- and word-level metrics primarily benefit from the phonetic subtask, whereas utterance-level metrics depend on both phonetic and prosodic subtasks, with prosodic features aiding stress and prosody. Besides, unlike \cite{fu23_interspeech}’s phoneme duration and phoneme subtasks, which underperform, our $S_{V}$, $S_{A}$, and $S_{F}$ subtasks yield superior results.

\begin{table}[t!]
\centering
\caption{The efficacy of only-phonetic (phn.) and only-prosodic (pros.) groups over phoneme and the averaged (Avg.) PCC in word and utterance.}
\label{tab:experimental_results.phonetic_prosodic_comparison}
\scalebox{0.97}{
\begin{tabularx}{\linewidth}{@{}>{\raggedright\arraybackslash}m{0.5cm}|@{\hskip 3pt}>{\raggedright\arraybackslash}m{2cm}|@{\hskip 2pt}>{\centering\arraybackslash}m{1.1cm}@{\hskip 2pt}>{\centering\arraybackslash}m{1cm}@{\hskip 2pt}*{2}{>{\centering\arraybackslash}X@{\hskip 2pt}}@{}}
\toprule
\multicolumn{2}{@{}>{\centering\arraybackslash}p{2.5cm}|}{\multirow{2}{*}{\textbf{Methods}}} & \multicolumn{2}{c}{\textbf{Phoneme}} & \multicolumn{1}{c}{\textbf{Word}} & \multicolumn{1}{c}{\textbf{Utterance}} \\
\cmidrule(lr){3-4} \cmidrule(lr){5-5} \cmidrule(lr){6-6}
\multicolumn{2}{c|}{} & MSE $\downarrow$ & PCC $\uparrow$ & Avg. PCC $\uparrow$ & Avg. PCC $\uparrow$ \\
\midrule
\multicolumn{2}{c|}{\makecell{HCBbf}} & \makecell{0.080\\(0.000)} & \makecell{0.651\\(0.003)} & \makecell{0.539\\(0.014)} & \makecell{0.779\\(0.025)} \\ \midrule 
\multirow{6}{*}{\footnotesize phn.} & \makecell[l]{\footnotesize +$S_P$} & \makecell{\underline{0.077}\\(0.001)} & \makecell{\textbf{0.670}\\(0.004)} & \makecell{\textbf{0.560}\\(0.015)} & \makecell{\textbf{0.804}\\(0.024)} \\
& \makecell[l]{\footnotesize +$S_P$$\otimes$$S_V$} & \makecell{\textbf{0.076}\\(0.000)} & \makecell{\textbf{0.670}\\(0.001)} & \makecell{\underline{0.545}\\(0.011)} & \makecell{0.767\\(0.012)} \\
& \makecell[l]{\footnotesize +$S_P$$\otimes$$S_V$$\otimes$$S_A$} & \makecell{\underline{0.077}\\(0.001)} & \makecell{\underline{0.664}\\(0.001)} & \makecell{0.527\\(0.019)} & \makecell{\underline{0.786}\\(0.030)} \\ \midrule \midrule
\multirow{6}{*}{\footnotesize pros.} &  \makecell[l]{\footnotesize +$S_D$} & \makecell{0.081\\(0.001)} & \makecell{0.647\\(0.005)} & \makecell{0.525\\(0.016)} & \makecell{0.768\\(0.027)} \\
& \makecell[l]{\footnotesize +$S_D$$\otimes$$S_F$} & \makecell{\underline{0.080}\\(0.001)} & \makecell{0.648\\(0.003)} & \makecell{0.525\\(0.012)} & \makecell{0.773\\(0.031)} \\
& \makecell[l]{\footnotesize +$S_D$$\otimes$$S_F$$\otimes$$S_R$} & \makecell{0.081\\(0.001)} & \makecell{0.641\\(0.007)} & \makecell{0.531\\(0.035)} & \makecell{0.757\\(0.037)} \\
\bottomrule
\end{tabularx}
}
\end{table}

\begin{table}[t!]
\centering
\caption{Comparison with \texttt{ER} \cite{do24_interspeech} and \texttt{HCF}. \texttt{Phn. Only}, \texttt{Wrd. Only} and \texttt{Utt. Only} indicates HCFs appending only on the APA phone, word or utterance regressors, respectively.}
\label{tab:experimental_results.fusing_handcrafted_features}
\scalebox{0.96}{
\begin{tabular}{l|cc|c|c}
\toprule
\multirow{2}{*}{\textbf{Methods}} & \multicolumn{2}{c}{\textbf{Phoneme}} & \multicolumn{1}{c}{\textbf{Word}} & \multicolumn{1}{c}{\textbf{Utterance}} \\
\cmidrule(lr){2-3} \cmidrule(lr){4-4} \cmidrule(lr){5-5}
 & MSE $\downarrow$ & PCC $\uparrow$ & Avg. PCC $\uparrow$ & Avg. PCC $\uparrow$ \\
\midrule
\makecell{HCBbf} & \makecell{0.080\\(0.000)} & \makecell{0.651\\(0.003)} & \makecell{0.539\\(0.014)} & \makecell{0.779\\(0.025)} \\
\makecell{\texttt{+ER} (All Heads)\cite{do24_interspeech}} & \makecell{0.081\\(0.001)} & \makecell{0.645\\(0.003)} & \makecell{0.519\\(0.037)} & \makecell{\underline{0.781}\\(0.046)} \\ 
\makecell{\texttt{+ER} (Phn. Only)} & \makecell{0.080\\(0.001)} & \makecell{0.649\\(0.002)} & \makecell{0.530\\(0.011)} & \makecell{0.759\\(0.015)} \\
\makecell{\texttt{+ER} (Wrd. Only)} & \makecell{\underline{0.078}\\(0.001)} & \makecell{\underline{0.663}\\(0.002)} & \makecell{\underline{0.559}\\(0.015)} & \makecell{0.757\\(0.028)} \\
\makecell{\texttt{+ER} (Utt. Only)} & \makecell{\underline{0.078}\\(0.001)} & \makecell{\underline{0.660}\\(0.003)} & \makecell{0.528\\(0.025)} & \makecell{0.770\\(0.039)} \\ \midrule
\makecell{\texttt{(1)+ER} (Utt. Only)} & \makecell{\textbf{0.076}\\(0.000)} & \makecell{\textbf{0.674}\\(0.001)} & \makecell{\textbf{0.548}\\(0.015)} & \makecell{0.738\\(0.042)} \\
\makecell{\texttt{(2)+ER} (Utt. Only)} & \makecell{\underline{0.078}\\(0.001)} & \makecell{0.659\\(0.002)} & \makecell{0.526\\(0.012)} & \makecell{0.752\\(0.058)} \\
\midrule \midrule
\makecell{\texttt{+HCF} (All Heads)} &  \makecell{0.084\\(0.001)} & \makecell{0.629\\(0.001)} & \makecell{0.538\\(0.016)} & \makecell{\underline{0.783}\\(0.014)} \\
\makecell{\texttt{+HCF} (Phn. Only)} & \makecell{0.080\\(0.001)} & \makecell{0.650\\(0.004)} & \makecell{0.537\\(0.013)} & \makecell{0.768\\(0.030)} \\
\makecell{\texttt{+HCF} (Wrd. Only)} & \makecell{0.080\\(0.001)} & \makecell{0.655\\(0.002)} & \makecell{0.529\\(0.026)} & \makecell{0.769\\(0.029)} \\
\makecell{\texttt{+HCF} (Utt. Only)} & \makecell{0.079\\(0.001)} & \makecell{\underline{0.658}\\(0.003)} & \makecell{0.527\\(0.024)} & \makecell{\underline{0.793}\\(0.010)} \\
\midrule
\makecell{\texttt{(1)+HCF} (Utt. Only)} & \makecell{\underline{0.077}\\(0.003)} & \makecell{\underline{0.670}\\(0.009)} & \makecell{\underline{0.540}\\(0.016)} & \makecell{\textbf{0.795}\\(0.015)} \\
\makecell{\texttt{(2)+HCF} (Utt. Only)} & \makecell{0.079\\(0.002)} & \makecell{\underline{0.656}\\(0.005)} & \makecell{0.537\\(0.015)} & \makecell{\underline{0.784}\\(0.023)} \\
\bottomrule
\end{tabular}
}
\vspace{-0.6cm}
\end{table}

We also pretrained on only-phonetic and only-prosodic groups to observe the efficacy of each task group. Table \ref{tab:experimental_results.phonetic_prosodic_comparison} shows that the phonetic group contributes more than the prosodic group does. Despite this, fluency and prosodic in the utterance-level get their peak performance when adding the $S_{F}$ subtask. This indicates the importance of both groups of subtasks in MTP.

\subsection{Fusing HCFs and Discussions}
\label{subsec:fusing_handcrafted_features_and_discussions}

Fusing HCFs into utterance-level decisions preserves APA model interpretability while enhancing holistic proficiency assessment. We compare two configurations from Table \ref{tab:experimental_results.without_dataaug}: HCBbf+\texttt{$S_{P}$} (denoted (1)) and HCBbf+\texttt{$S_{P} \otimes S_{D} \otimes S_{V}  \otimes S_{A} \otimes S_{F}$} (denoted (2)), representing for the phone-word and utterance groups, respectively. In Table \ref{tab:experimental_results.fusing_handcrafted_features}, we first compare our approach with \cite{do24_interspeech}, adding ER metrics, which introduce the character error rate (CER) and phoneme match error rate (MER-P), improves the utterance-level PCC from 0.779 to 0.781. However, applying ER fusion solely at utterance-level regressors (Utt. Only) outperforms fusion across all regressors (All Heads), as well as phone-only (Phn. Only) and word-only (Wrd. Only) configurations, making it our preferred strategy for further comparisons.

\begin{figure}[t!]
  \centering
  \includegraphics[width=\linewidth]{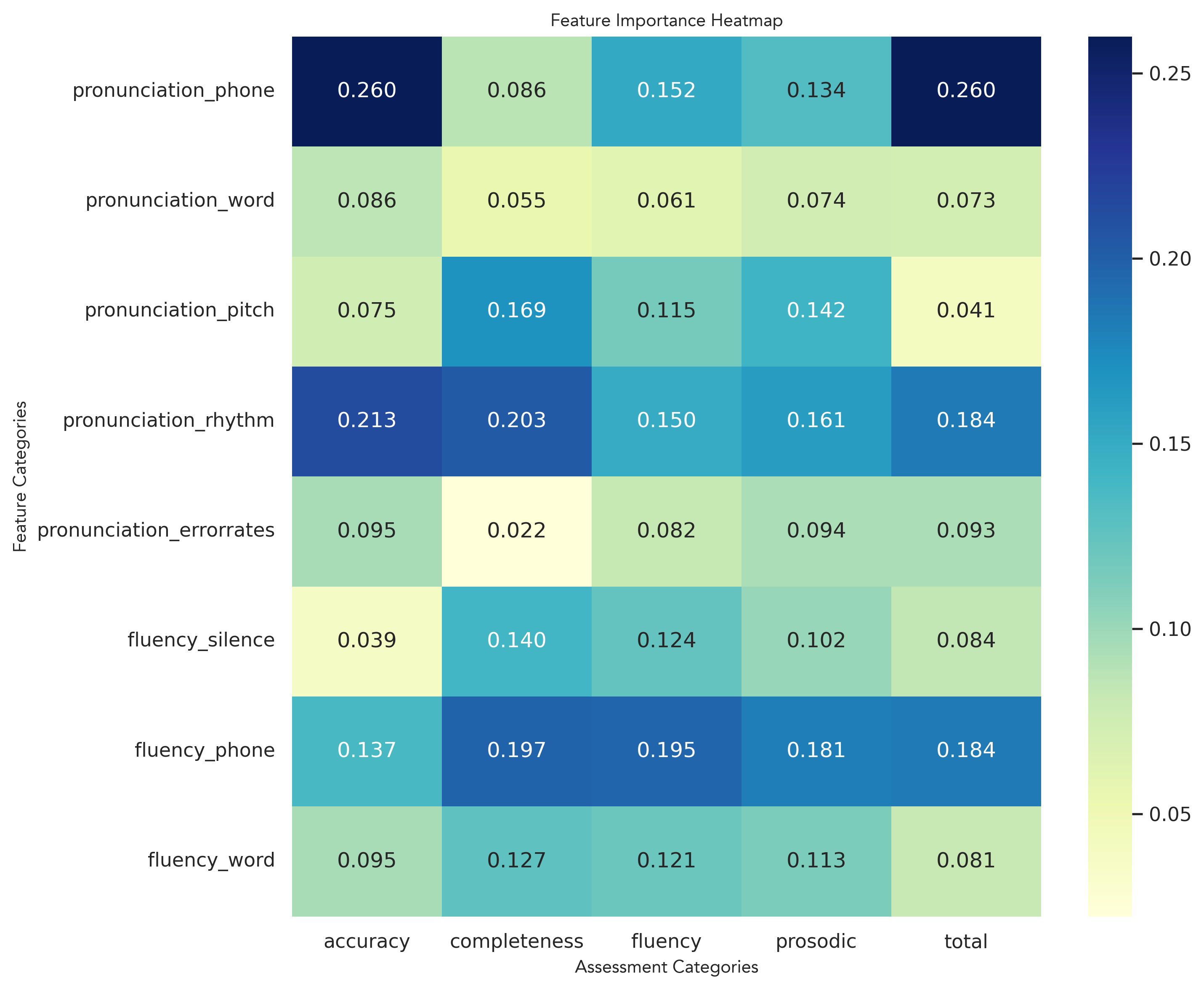}
  \caption{The illustration of the relation in assessment categories (the aspects in the utterance-level) and feature categories (the features column in Table \ref{tab:asa_attributes}).}
  \label{fig:heatmap}
  \vspace{-0.5cm}
\end{figure}

Then, we integrate highly relevant HCFs into the final decisions. Figure \ref{fig:heatmap} illustrates HCF importance. Surprisingly, \texttt{ER} contributes less than expected, while with \texttt{fluency\_phone} and \texttt{pronunciation\_phone} scoring above 0.150, highlighting the significance of phonetic features, \texttt{pronunciation\_rhythm} contributes across multiple aspects. \texttt{+HFC} Phn. Only and Wrd. Only do not outperform Utt. Only at the utterance level. And \texttt{(1)+HCF} (Utt. Only) achieves a PCC of 0.795. This suggests that HCFs enhance utterance-level ASA more than phoneme/word-level APA, ideal for holistic evaluation in CALL systems, while phoneme- and word-level evaluations benefit more from fine-grained features.

To bridge APA and ASA, we further discuss how APA’s phonetic and prosodic outputs contribute to holistic speaking proficiency. Correlation analysis reveals that APA’s phoneme accuracy and stress alignment strongly influence ASA’s fluency scores (e.g., speech rate, silence duration), as well as ASA's rhythm scores, which represent 
prosodic variations, making it more vivid and natural through elements. Regular prosodic changes help speakers maintain stable speech rates while making it easier for listeners to understand meaning, validating the integration of HCFs. These HCFs, derived from human-designed formulas, produce transparent feedback, such as slow speech rate or excessive pauses, enabling learners to address specific issues and educators to design targeted exercises. 

\section*{Conclusions}

This work proposes a joint modelling approach for phonetic and prosodic subtasks during the pretraining of the APA model's encoder, enhancing its efficacy. Additionally, we explore HCFs, bridging the gap toward a comprehensive system for holistic assessment, offering personalized feedback for L2 learners and data-driven insights for educators.












\bibliographystyle{IEEEtran}
\bibliography{mybib}

\end{document}